\documentclass{article}

\usepackage{microtype}
\usepackage{graphicx}
\usepackage{subcaption}
\usepackage{booktabs} 

\usepackage{hyperref}

\usepackage[preprint]{icml2026}

\usepackage{amsmath}
\usepackage{amssymb}
\usepackage{mathtools}
\usepackage{amsthm}

\usepackage{enumitem}
\usepackage{multirow}
\usepackage{booktabs}
\usepackage[table]{xcolor}
\usepackage{amsmath}
\usepackage{xurl}

\usepackage[capitalize,noabbrev]{cleveref}

\theoremstyle{plain}

\theoremstyle{definition}

\theoremstyle{remark}

\usepackage[textsize=tiny]{todonotes}

\icmltitlerunning{DynSplit-KV: Dynamic Semantic Splitting for KVCache Compression in Efficient Long-Context LLM Inference}

\begin{document}


\twocolumn[
  \icmltitle{DynSplit-KV: Dynamic Semantic Splitting for  KVCache Compression in Efficient Long-Context LLM Inference}


  \icmlsetsymbol{equal}{*}
  \icmlsetsymbol{Correspond}{$^\dagger$}

  \begin{icmlauthorlist}
    \icmlauthor{Jiancai Ye}{equal,SJTU}
    \icmlauthor{Jun Liu}{equal,SJTU}
    \icmlauthor{Qingchen Li}{SJTU}
    \icmlauthor{Tianlang Zhao}{SJTU}
    \icmlauthor{Hanbin Zhang}{SJTU}
    \icmlauthor{Jiayi Pan}{SJTU}
    \icmlauthor{Ningyi Xu}{SJTU}
    \icmlauthor{Guohao Dai}{Correspond,SJTU,Infinigence-AI,SII}
  \end{icmlauthorlist}

  \icmlaffiliation{SJTU}{Shanghai Jiao Tong University, Shanghai, China}
  \icmlaffiliation{Infinigence-AI}{Infinigence-AI, Shanghai, China}
  \icmlaffiliation{SII}{SII, Shanghai, China}

  \icmlcorrespondingauthor{Guohao Dai}{daiguohao@sjtu.edu.cn}

  

  \vskip 0.1in
  
]



\printAffiliationsAndNotice{}  

\begin{abstract}

Although Key-Value (KV) Cache is essential for efficient large language models (LLMs) inference, its growing memory footprint in long-context scenarios poses a significant bottleneck, making KVCache compression crucial. Current compression methods rely on \textbf{rigid splitting} strategies, such as fixed intervals or pre-defined delimiters. We observe that rigid splitting suffers from significant accuracy degradation (ranging from 5.5\% to 55.1\%) across different scenarios, owing to the scenario-dependent nature of the semantic boundaries. This highlights the necessity of \textbf{dynamic semantic splitting} to match semantics. To achieve this, we face two challenges. (1) \textbf{Improper delimiter selection} misaligns semantics with the KVCache, resulting in 28.6\% accuracy loss. (2) \textbf{Variable-length} blocks after splitting introduce over 73.1\% additional inference overhead.
To address the above challenges, we propose \textit{DynSplit-KV}, a KVCache compression method that dynamically identifies delimiters for splitting. We propose: (1) a dynamic importance-aware delimiter selection strategy, improving accuracy by 49.9\%. (2) A uniform mapping strategy that transforms variable-length semantic blocks into a fixed-length format, reducing inference overhead by 4.9$\times$. Experiments show that \textit{DynSplit-KV} achieves the highest accuracy, 2.2$\times$ speedup compared with FlashAttention and 2.6$\times$ peak memory reduction in long-context scenarios.

\begin{figure}[!t]
  \centering
  \includegraphics[width=0.49\textwidth]{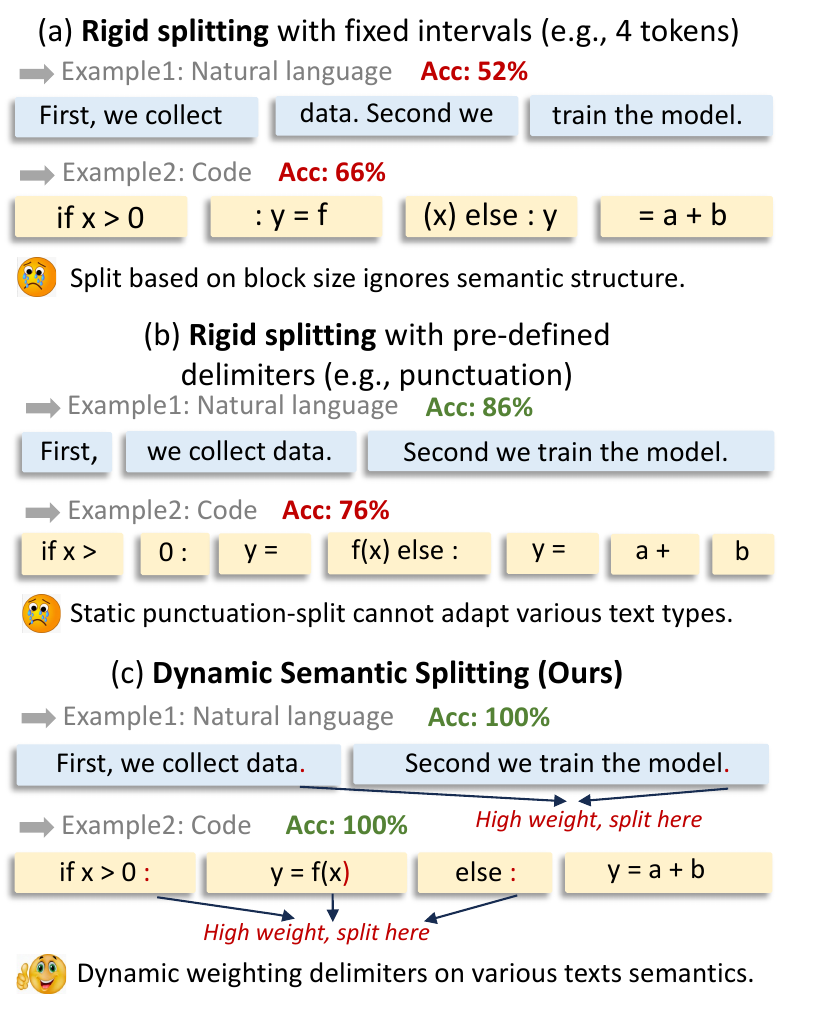}
  \vskip 0.1in
  \setlength{\abovecaptionskip}{4pt} 
  \caption{Comparison summary: (a) Rigid splitting with fixed intervals (e.g., InfLLM~\citep{xiao2024infllm}, Quest~\citep{tang2024quest}, ChunkKV~\citep{liu2025chunkkv}) (b) Rigid splitting with pre-defined delimiters (e.g., SentenceKV~\citep{zhu2025sentencekv}) (c) Ours: Dynamic semantic splitting by weighting the delimiters dynamically. Note: Accuracy data are obtained from experimental evaluations of InfLLM, SentenceKV, and our method on datasets of LongBench~\citep{bai2023longbench}.  }
  \vskip -0.2in
  \label{fig: comparison summary}
\end{figure}

\end{abstract}

\section{Introduction}

\begin{figure}[!t]
  \centering
  \includegraphics[width=0.49\textwidth]{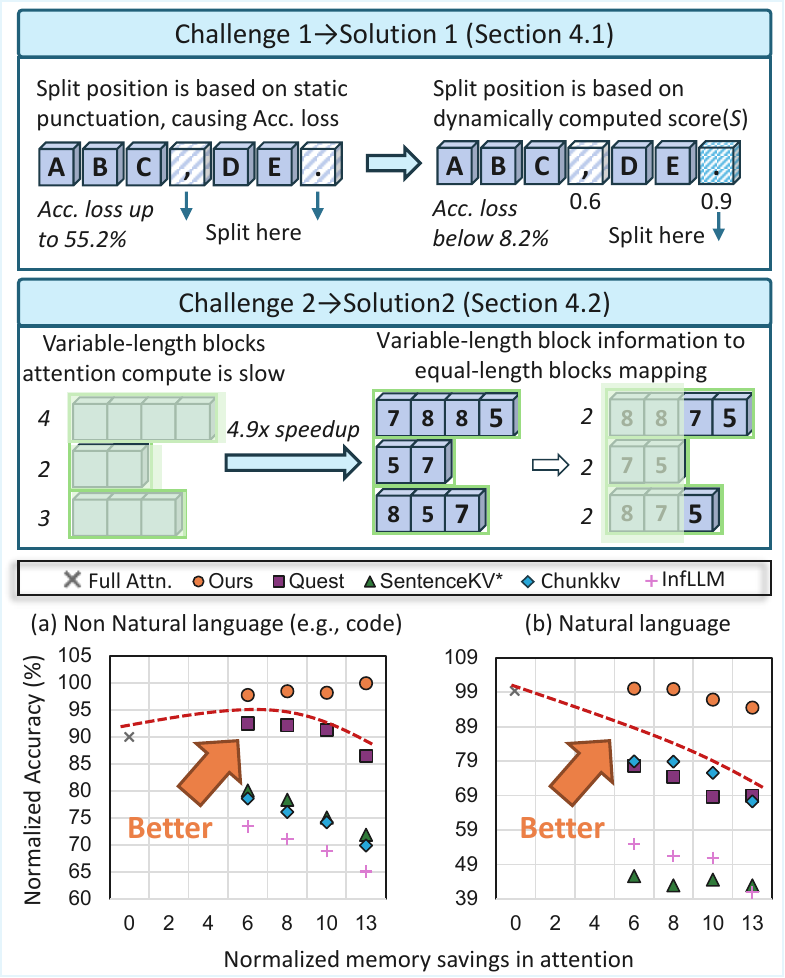}
  \vskip 0.1in
  \setlength{\abovecaptionskip}{4pt} 
  \caption{Challenges, solutions, and performance summary. Note: Accuracy data are obtained from experimental evaluations of baselines and our method on datasets of LongBench~\citep{bai2023longbench}.}
 \vskip -0.2in
  \label{fig: challenge overview}
\end{figure}

Large language models (LLMs) have demonstrated their abilities to process long contexts, with recent models supporting up to 1M-10M tokens~\citep{anthropic2025claude37, comanici2025gemini}. This facilitates the development of long-context applications, such as code generation~\citep{qwen3coder2025, claudecoder2025} and multi-document question answering~\cite{openai2025deepresearch, anthropic2025research} in the agent era. However, efficient long-context inference faces challenges related to KVCache, which stores previous KV activations to avoid redundant computation. As the KVCache grows with sequence length, its increasing memory footprint and per-token KVCache accesses during token generation lead to low inference throughput. For example, the KVCache generated by the Llama2-13B model with 128k context length fills 80GB of memory (NVIDIA A100 has only 80GB of memory), causing the KVCache access to account for over 80.0\% of the total inference latency, which is detailed described in the Appendix~\ref{A_1}. Consequently, compressing the KVCache has become an effective approach to mitigate this memory bottleneck, reducing GPU memory usage and accelerating inference. KVCache compression approaches are generally divided into two stages: KVCache compression, which splits and compresses the input or newly generated KVCache, and KVCache selection, which selects the minimal KVCache subset to generate new tokens while meeting accuracy requirements.

\textbf{Existing KVCache compression approaches are based on rigid splitting strategies} with either fixed intervals or pre-defined delimiters~\citep {xiao2023efficient,li2024snapkv,liu2024clusterkv,tang2024quest,zhang2024pqcache}, as illustrated in Figure~\ref{fig: comparison summary}. Rigid splitting with fixed intervals segments the KVCache to fixed-length semantic blocks to perform compression, e.g., eviction or sparsification, improving the efficiency but ignoring semantic correlations among different blocks. 2. Rigid splitting with pre-defined delimiters partitions the KVCache using pre-defined delimiters (e.g., punctuations) to preserve semantic structure, but these rely on natural language conventions and fail on other text types. Overall, these rigid splitting strategies miss semantic boundaries, as they rely on fixed intervals or pre-defined delimiters for KVCache splitting.

\textbf{We observe that semantic boundaries are scenario-dependent, highlighting the need for dynamic semantic splitting to match semantics.} Rigid splitting strategies suffers from significant accuracy degradation (ranging from 5.5\% to 55.1\%) across different scenarios. A good delimiter should mark semantic boundaries while enabling preceding content to generate subsequent content. To quantify delimiter weights, we adopt the attention score between different KVCache segments, capturing their dependencies. Our results show that across scenarios and models, the importance of the same delimiter varies significantly, 45\% between code and natural language, and 24\% across models (Section~\ref{sec: observations}).

However, achieving this observation faces two challenges, which are caused by the dynamic nature of the semantic structure and block length. Challenge 1: \textbf{Improper delimiter selection} misaligns semantic boundaries with the KVCache structure, causing semantically related tokens to be split across blocks and resulting in a 28.6\% accuracy loss. Challenge 2: \textbf{Variable-length} blocks after splitting introduce over 73.1\% additional time overhead during KV selection, as block-wise scoring and selection can no longer be efficiently parallelized and require extra handling for irregular block sizes.

To address these challenges, we propose DynSplit-KV, a KVCache compression method that dynamically identifies delimiters for semantic splitting. The contributions are as follows:

\begin{itemize}
    \item We make the key observation that different models and contexts exhibit varying preferences for delimiters, indicating that static KVCache splitting is insufficient. Based on this observation, we propose DynSplit-KV.

    \item In the compression stage, we propose DD-Select, a dynamic importance-aware delimiter selection strategy by scoring delimiters based on their importance weights and the lengths between adjacent delimiters, improving accuracy by 49.9\% (Section~\ref{sec: method1}).

    \item In the selection stage, we design V2F, a method to map variable-length blocks to fixed-length blocks by using block scores to represent token scores within each block, reducing extra time overhead by 4.9$\times$ (Section~\ref{sec: method2}).

\end{itemize}

Experiments show that \textit{DynSplit-KV} achieves the highest accuracy, 2.2$\times$ speedup compared with FlashAttention and 2.6$\times$ peak memory reduction in long-context scenarios (Section~\ref{sec: experiment}).

\section{Related Works}

\subsection{Rigid Splitting with Fixed Intervals}
Rigid splitting with fixed intervals mainly consists of two primary categories: token-level and block-level.

\textbf{Token-level.} To reduce the memory footprint, token-level approaches keep only the most critical token KV pairs while discarding unnecessary tokens.
For example, StreamingLLM~\citep{xiao2023efficient} retains attention sinks and recent KV pairs to address the limitations of windowed attention. SnapKV~\citep{li2024snapkv} selects important tokens for future generations based on local prompt windows. H2O~\citep{zhang2023h2o} introduces a low-overhead token eviction mechanism that leverages cumulative attention scores to maintain the KVCache. These methods operate at the token level, ignoring sequential semantics, resulting in a 10\%–30\% drop in accuracy. Besides, several methods have been proposed to optimize KVCache quantization~\cite{liu2024kivi, hooper2024kvquant, xiao2023smoothquant}, which reduces token bit width and is orthogonal to our approach.

\textbf{Block-level.} Block-level approaches organize KVCache into blocks or pages and perform dynamic KV selection during inference to reduce memory access overhead. Quest~\citep{tang2024quest} segments tokens into fixed-size pages and selects pages by approximating the maximum attention score within each page. ClusterKV~\citep{liu2024clusterkv} clusters KV pairs based on semantic similarity and dynamically loads relevant clusters according to query–cluster relevance. InfLLM~\citep{xiao2024infllm} partitions the KV cache into fixed-length blocks and dynamically loads relevant blocks during inference based on approximate attention estimation. These approaches still disrupt continuous semantic structure. There are also methods like ShadowKV~\cite{sun2024shadowkv} that compress the KVCache via low-rank decomposition, which are orthogonal to our dynamic segmentation approach.

\subsection{Rigid Splitting with Pre-defined Delimiters}
These studies manage KVCache at the sentence-level to better preserve semantic coherence.
SentenceKV~\cite{zhu2025sentencekv} splits tokens into sentence-level semantic units by predefined punctuations during prefill, stores compact sentence representations on GPU, and selectively retrieves relevant sentence KVs during decoding based on query–sentence similarity. SABlock~\cite{chen2025sablock} aligns cache boundaries with semantic segments and adaptively determines block sizes under a fixed cache budget to improve compression efficiency with preserved semantic integrity. Despite improved semantics, these methods rely on predefined delimiters and incur additional preprocessing or retrieval overhead due to variable-length semantic blocks, which limits their adaptability in real-time inference.

\section{Observations}
\label{sec: observations}

\begin{figure*}[!t]  
  \centering
  \includegraphics[width=0.98\textwidth]{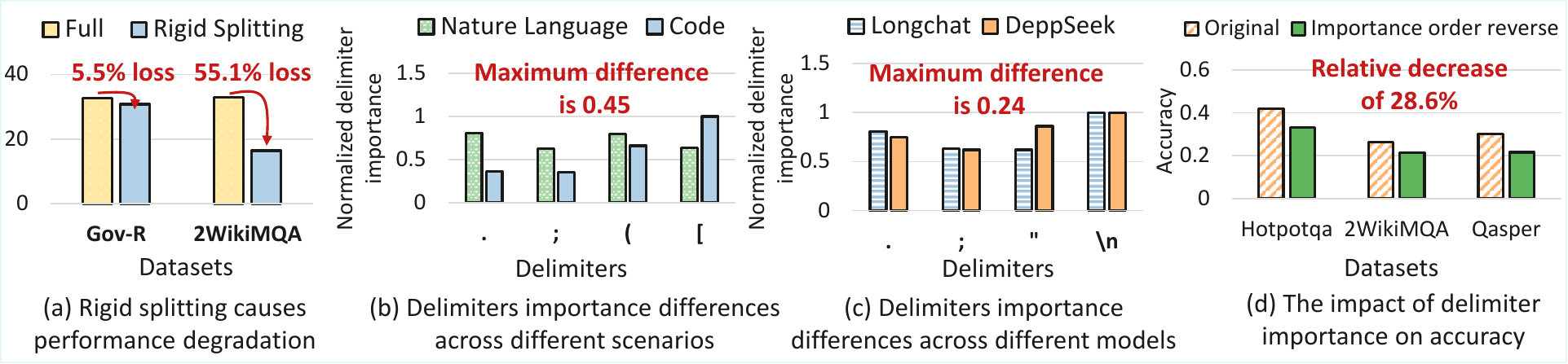}
  \setlength{\abovecaptionskip}{4pt} 
  \caption{Observation. (a) Impacts of rigid splitting and delimiter importance (across (b) scenarios/ (c) models, (d) on accuracy).}
  \label{fig: observation}
\end{figure*}

\begin{algorithm}[t]
\caption{Delimiter Importance Scoring via Attention Dependency (Section~\ref{sec: observations})}
\label{alg: delimiter}
\begin{algorithmic}[1]
\REQUIRE Token sequence $T$, transformer model $M$, candidate delimiter positions $\mathcal{D}$, overlap size $R$, future window size $W$
\ENSURE Importance score $s_i$ for each $i \in \mathcal{D}$
\STATE Run $M$ on $T$ and collect attention maps $\{A^{(l,h)}\}$
\FOR{each delimiter position $i \in \mathcal{D}$}
    \STATE $\mathcal{F}_i \leftarrow \{i+1,\ldots,i+W\}$
    \STATE $\mathcal{O}_i \leftarrow \{\max(0,i-R+1),\ldots,i\}$
    \STATE $\mathcal{D}_i \leftarrow \{0,\ldots,\max(0,i-R)\}$
    \STATE $\text{OverlapCBD}_i \leftarrow \mathbb{E}_{l,h,q \in \mathcal{F}_i} \sum_{k \in \mathcal{O}_i} A^{(l,h)}_{q,k}$
    \STATE $\text{DropCBD}_i \leftarrow \mathbb{E}_{l,h,q \in \mathcal{F}_i} \sum_{k \in \mathcal{D}_i} A^{(l,h)}_{q,k}$
    \STATE $s_i \leftarrow \text{OverlapCBD}_i - \alpha \cdot \text{DropCBD}_i$
\ENDFOR
\STATE \textbf{return} $\{s_i\}_{i \in \mathcal{D}}$

\end{algorithmic}
\end{algorithm}

\textbf{Limitations of Rigid Splitting.} We observe that rigid splitting suffers from significant accuracy degradation across different scenarios, with relative drops ranging from 5.5\% to 55.1\%. This performance loss arises because semantic boundaries are scenario-dependent, and fixed splitting strategies fail to align with the underlying semantic structure, as shown in Figure~\ref{fig: observation} (a).

\textbf{Delimiter importance estimation via attention dependency.} We believe that a good delimiter should mark semantic boundaries.
To verify our conjecture, we define a delimiter importance score by measuring how future tokens attend to recently retained context versus distant discarded context. This attention-based formulation captures a semantically valid boundary, where a good delimiter preserves local semantic coherence while minimizing long-range dependency. Formally, for each candidate delimiter position $i$, we partition the preceding context into a retained region $\mathcal{O}_i$ and a discarded region $\mathcal{D}_i$, and compute the attention of tokens in a future window $\mathcal{F}_i$. The importance score $s_i$ is:
\[
s_i = 
\mathbb{E}_{l,h,q \in \mathcal{F}_i}
\left(
\sum_{k \in \mathcal{O}_i} A^{(l,h)}_{q,k}
-
\alpha \sum_{k \in \mathcal{D}_i} A^{(l,h)}_{q,k}
\right),
\]
Where $A^{(l,h)}$ is the attention map for head $h$ in layer $l$, and $\alpha$ penalizes long-range dependencies. High scores indicate delimiters that preserve local semantic coherence while minimizing reliance on distant tokens, providing a reliable criterion for dynamic KVCache splitting. The importance of the same delimiter can vary significantly across different inference scenarios and models, with up to 54.5\% difference between code and natural language, and 49.1\% across models, detailed in Figure~\ref{fig: observation}. Typically, $W$ is set to 8, $R$ to 128, and $\alpha$ to 1, as detailed in Algorithm~\ref{alg: delimiter}.

\textbf{Effect of delimiter importance on inference accuracy.} Delimiter importance can be stably determined during the prefill stage, with scores converging within 0.1 using only a few tokens, demonstrating their potential for real-time dynamic inference. We conduct a controlled experiment to examine the effect of delimiter importance during inference. Reversing the estimated importance order while keeping all other settings identical results in an accuracy drop of up to 28.6\% on some datasets, as shown in Figure~\ref{fig: observation} (d). This demonstrates that the relative ordering of delimiter importance is critical to model performance, motivating the exploration of dynamic semantic splitting strategies that adapt delimiter selection to different contexts. An example of detail delimiter importance is shown in Appendix~Table~\ref{tab: delimiter importance example}.

\section{DynSplit-KV}
\label{sec: method}

\subsection{\underline{D}ynamic Importance-aware \underline{D}elimiter \underline{Select}ion Strategy (DD-Select)} 

\label{sec: method1}

\begin{figure}[!t]
  \centering
  \includegraphics[width=0.49\textwidth]{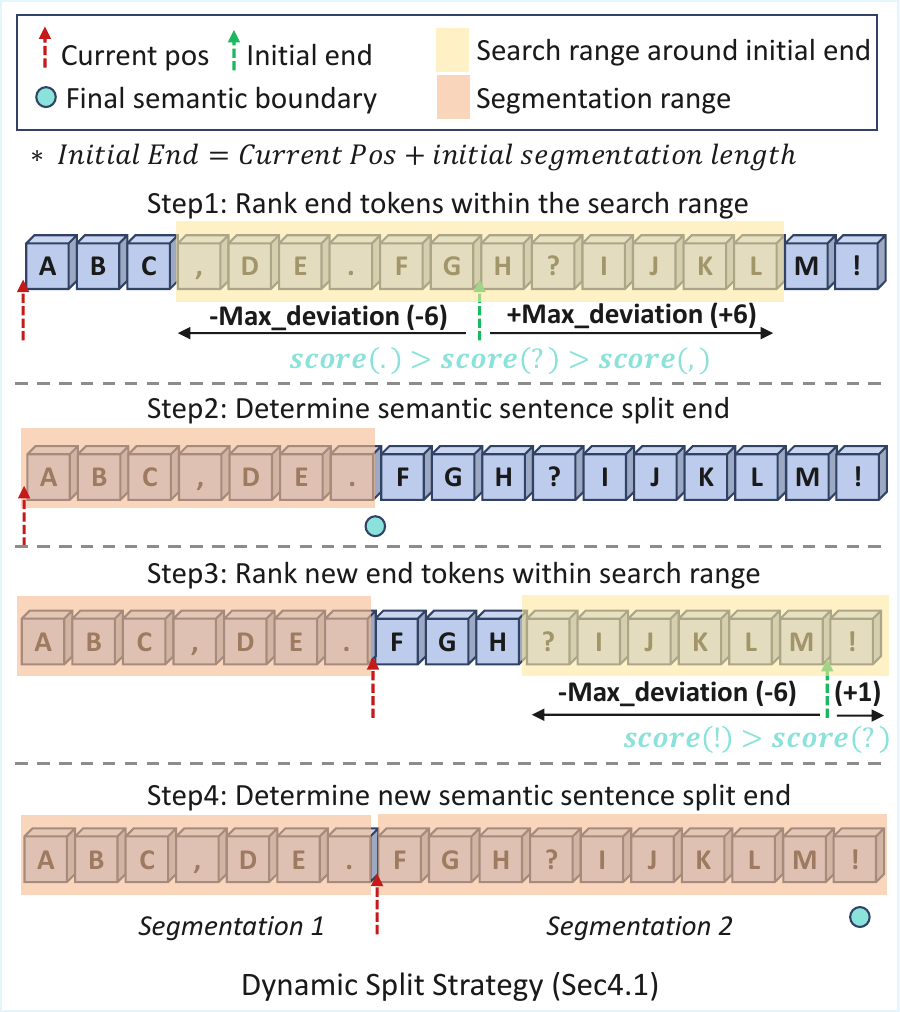}
  \vspace{-10pt}
  \setlength{\abovecaptionskip}{4pt} 
  \caption{DD-Select Pipeline.}
  \vskip -0.2in
  \label{fig: method1}
\end{figure}

We propose a dynamic importance-aware delimiter selection strategy that segments input sequences into semantically coherent blocks while regulating segment length. DD-Select balances semantic preservation and computational efficiency, see Figure~\ref{fig: method1}.

\textbf{Semantic Boundary Tokens.} We define a set of \emph{semantic boundary tokens} (e.g., punctuation, newlines) and assign each an importance weight (Detailed in Section~\ref{sec: observations}) reflecting its likelihood of marking a semantic break. Let $\mathcal{B} = \{b_i\}$ denote the set of boundary tokens and $w_i \in [0,1]$ their weights. Tokens with higher weights are considered more likely to indicate semantic boundaries.

\textbf{Dynamic Segmentation Procedure.} Let $x_{1:L}$ denote the input sequence of length $L$, and $C$ the desired base chunk size. Chunks are determined iteratively as follows:

\begin{enumerate}[itemsep=0pt, topsep=0pt, partopsep=0pt, parsep=0pt, leftmargin=*]
    \item \textbf{Current pos ($s_c$) and Initial end ($s_e$):} The starting position of the current chunk and the ideal end of the chunk ignoring semantic boundaries. Here $s_e=s_c+C$.
    \item \textbf{Search range around initial end:} window $[s_e-\Delta,\, s_e+\Delta]$, where $\Delta$ is the maximum allowed deviation.
    \item \textbf{Final semantic boundary ($e^*$):} among the semantic boundaries in the search range, select the one that maximizes a combined score of semantic importance and proximity:
    \[
    e^* = \arg\max_{e \in [s_e-\Delta,\, s_e+\Delta]} \big( \alpha \, w_e + (1-\alpha) \, p_e \big),
    \]
    where $w_e$ is the semantic weight of delimiter $e$, $p_e$ measures how close $e$ is to the ideal chunk end $s_e$, and $\alpha \in [0,1]$ balances semantic importance and length regularization.
    \item \textbf{Segmentation range:} define the chunk as $[s, e^*)$, then update current pos to $e^*$ for the next iteration.
\end{enumerate}

\textbf{Incremental Update during Decoding.} During autoregressive decoding, previously computed chunks are cached. When new tokens arrive, only the most recent segmentation ranges are recomputed. This dynamic update allows chunks to adapt to evolving contexts.

\begin{figure*}[!tp]
    \centering
    \includegraphics[width=0.99\textwidth]{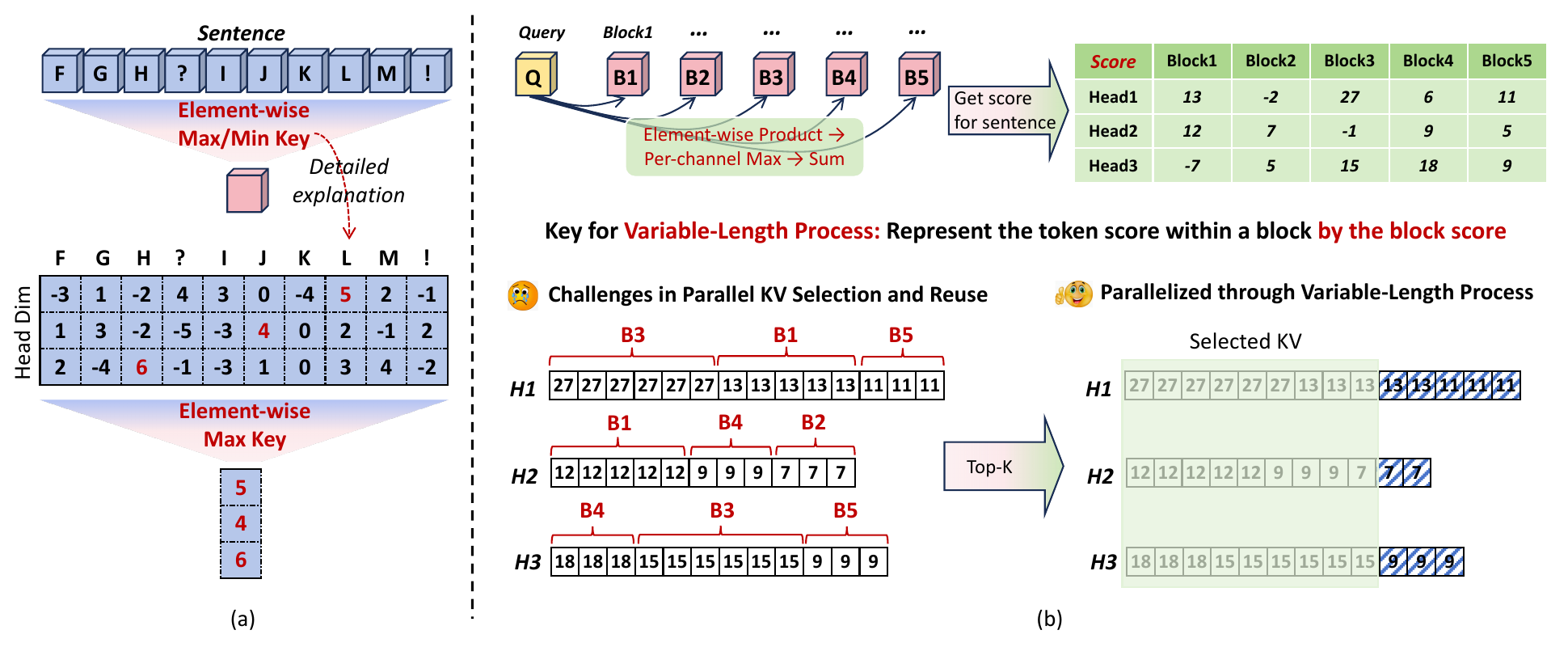}
    \vspace{-5pt}
    \caption{(a) Compression Strategy. (b) Variable-to-Fixed Block Mapping Strategy.}
    \vspace{-10pt}
    \label{fig: method2}
\end{figure*}

\subsection{\underline{V}ariable-\underline{to}-\underline{F}ixed Block Mapping Strategy (V2F)} 
\label{sec: method2}



We propose V2F, a strategy that maps variable-length semantic blocks to fixed-length representations for efficient KVCache attention, comprising KVCache compression, top-$k$ block selection, and block-to-token mapping (Figure~\ref{fig: method2}).

\textbf{KVCache Compression.} Given the variable-length semantic blocks $\{B_i\}$, we compress their key and value matrices using \emph{element-wise maximum and minimum} across the tokens in each block. This produces a fixed-length, memory-efficient representation per block. To show that the main advantages arise from the splitting strategy rather than the compression method, we further evaluate our approach with mean pooling–based compression (see Appendix~\ref{A_2}).

\textbf{Top-$k$ Block Selection.} We compute an importance score for each compressed block $\hat{B}_i$ using the query–compressed vector product, and select the top-$k$ highest-scoring blocks as attention candidates. The compressed representations enable efficient scoring for variable-length blocks without iterating over individual tokens.

\textbf{Block-to-Token Mapping.} Direct attention on variable-length blocks is inefficient due to differing token counts. We therefore map each block's importance score to all tokens within the block, producing token-level scores without explicit token-wise computation. Formally, if block $B_i$ has importance $s_i$, we assign $s_i$ to each token $t \in B_i$.  

Top-$k$ tokens are then selected based on these mapped scores, enabling parallel attention computation:
\begingroup
\setlength{\abovedisplayskip}{2pt}
\setlength{\belowdisplayskip}{2pt}
\[
\text{Top-k tokens} = \mathrm{topk}\Big(\{s_i \mid t \in B_i\}\Big)
\]
\endgroup

This mapping effectively translates variable-length block importance into token-level importance, allowing high-efficiency KVCache updates and attention calculation, while avoiding costly per-token scoring.

\textbf{Summary.} V2F unifies semantic-aware compression, top-$k$ block selection and block-to-token mapping, converting variable-length semantic blocks to fixed-length representations and guiding token selection via block scores to enable efficient parallel attention with preserved semantic fidelity. We integrate V2F with KVCache selection and KVCache reuse for practical inference (see Appendix~\ref{B_2}).

\section{Evaluation}
\label{sec: experiment}

In this section, we demonstrate the accuracy and efficiency of DynSplit-KV through extensive experiments.

\subsection{Evaluation Setup}

\begin{table*}[ht]
\centering
\scriptsize
\caption{Main accuracy results. In the three categories of related work, bold indicates the first rank in performance, while underlining indicates the second rank in performance.}
\label{tab: Longbench overview}
\begin{tabular}{c|c|cc|cc|cc|c|cc|>{\columncolor{gray!15}}c} 
\toprule[1.5pt]
& \multicolumn{1}{c|}{} 
  & \multicolumn{2}{c|}{Single-Document QA} 
  & \multicolumn{2}{c|}{Multi-Document QA} 
  & \multicolumn{2}{c|}{Summarization} 
  & Few-shot & \multicolumn{2}{c|}{Code}  & \cellcolor{gray!15} \\
\cmidrule(lr){3-4} \cmidrule(lr){5-6} \cmidrule(lr){7-8} \cmidrule(lr){9-9} \cmidrule(lr){10-11}
\multicolumn{1}{c|}{\begin{tabular}{c}KV Usage\\Rate = 0.1\end{tabular}}
& \multicolumn{1}{c|}{Method} 
& NarrativeQA & Qasper 
& HotpotQA & 2WikiMQA 
& Gov-R & MultiNews 
& TriviaQA & Lcc & R-P & \cellcolor{gray!15}\textbf{Avg.} \\ 
\midrule[1.2pt]
\multirow{2}{*}{Token-level} 
    & H2o          & 19.08 & 18.13 & 23.80 & 18.13 & 23.81 & 22.17 & 83.79 & 41.36 & 40.57   & \cellcolor{gray!15}32.32    \\
\cmidrule[0.6pt](l{2pt}r{2pt}){2-12}
    & StreamingLLM & 20.56 & 12.16 & 22.25 & 17.38 & 20.26 & 19.86 & \underline{85.06} & 42.18 & 44.68   & \cellcolor{gray!15}31.60     \\ 
\midrule[0.8pt]
\multirow{3}{*}{Block-level}
    & InfLLM  & 13.31 & 15.53 & 28.04 & 13.83 & 27.57 & 23.49 & 74.81 & 36.35 & 22.95   & \cellcolor{gray!15}28.43 \\ 
\cmidrule[0.6pt](l{2pt}r{2pt}){2-12}
    & Quest        & 20.87 & 22.93 & 32.90 & 18.65 & 30.67 & 25.51 & 78.47 & \underline{51.34} & \underline{56.16}   & \cellcolor{gray!15}\underline{37.50}    \\ 
\cmidrule[0.6pt](l{2pt}r{2pt}){2-12}
    & ChunkKV      & 20.55 & \underline{29.32} & \underline{37.40} & \underline{20.52} & \underline{32.75} & \textbf{26.54} & 78.53 & 39.51 & 44.68   & \cellcolor{gray!15}36.63    \\ 
\midrule[0.8pt]
\multirow{1}{*}{Sentence-level}  
    & SentenceKV*      & \underline{22.07} & 16.45 & 19.77 & 12.10 & 30.79 & 23.21 & 69.51 & 44.61 & 46.21   & \cellcolor{gray!15}31.64 \\
\midrule[0.8pt]
\rowcolor{cyan!15}
Ours & DynSplit-KV  & \textbf{25.47} & \textbf{30.20} & \textbf{42.01} & \textbf{26.26} & \textbf{33.15} & \underline{26.40} & \textbf{86.20} & \textbf{54.92} & \textbf{59.11}   & \cellcolor{gray!15}\textbf{42.64}    \\ 
\midrule[0.8pt]
Full KV & Full         & 26.47 & 32.91 & 43.72 & 26.97 & 32.59 & 26.97 & 85.74 & 55.18 & 53.94   & \cellcolor{gray!15}42.72    \\
\bottomrule[1.5pt]
\end{tabular}
\end{table*}

\textbf{Models and Benchmarks.} We used three representative models for our evaluation: the Mistral-7B-Instruct-v0.2~\citep{jiang2023mistral7b}, LongChat-v1.5-7b~\citep{li2023long}, DeepSeek R1 Distill Llama-8B~\citep{Guo_2025} and Llama3-8B-1M~\citep{gradientlongcontextllama3}. We evaluate our approach on three challenging long-context benchmarks: LongBench~\citep{bai2023longbench}, LongBench v2~\citep{bai2025longbench}, and the passkey retrieval task~\citep{peng2023yarn}, covering a wide range of long-context tasks with varying context lengths, including natural language and code tasks. The window length parameter between the initial end and the current position in DynSplit-KV is set to 14.

\textbf{Hardware Platforms.} Our experiments were conducted on a machine with eight NVIDIA A800 80GB GPUs and two Intel Xeon Platinum 8358 CPUs. For our method, DynSplit-KV, we consider two variants: one that keeps KVCache on the GPU for pure GPU inference, and another that offloads KVCache from the GPU to the CPU for CPU–GPU collaborative inference. These two settings respectively correspond to scenarios with sufficient GPU memory and limited GPU memory, covering typical long-context and ultra-long-context inference scenarios.

\textbf{Baselines.} We select representative methods from each of the three categories of related work as baselines for a fair comparison with DynSplit-KV. \textbf{Token-level approaches:} We choose StreamingLLM[ICLR'24]~\citep{xiao2023efficient}, the first work that proposes the concept of attention sink. We also choose H2O[NIPS'23]~\citep{zhang2023h2o}, which propose Heavy Hitter Oracle, a KVCache eviction policy. \textbf{Block-level approaches:} We select Quest[ICML'24]~\citep{tang2024quest}, a classical fixed-length block-based method, InfLLM[NIPS'24]~\citep{xiao2024infllm}, which partitions the KVCache into blocks via clustering and ChunkKV[NIPS'25]~\citep{liu2025chunkkv}, which is the current SOTA block-based KV method. \textbf{Sentence-level approaches:} We select SentenceKV[COLM'25]~\citep{zhu2025sentencekv}, a classical method that splits the KVCache based on punctuation.

\subsection{Accuracy Evaluation}
\label{sec: accuracy evaluation}

\textbf{LongBench.} To validate that DynSplit-KV outperforms baseline methods on general long-context datasets, we evaluate our method and the baselines on multiple datasets in LongBench. SentenceKV* indicates that we reproduced its KVCache splitting method while using the same compression approach as DynSplit-KV, to highlight the advantages of our splitting strategy. As shown in the table~\ref{tab: Longbench overview}, DynSplit-KV consistently outperforms all baselines across multiple datasets at the same KVCache usage. Token-level methods, which generally rely on KVCache eviction, achieve the lowest accuracy. Block-level methods perform better than token-level methods, while sentence-level methods outperform block-level methods on some datasets but perform poorly on others, with relative accuracy drops from 5.5\% to 55.1\%. This is because the distribution of punctuation varies dynamically across different text scenarios, and treating all punctuation equally as delimiters is not effective.

\begin{figure*}[!t]  
  \centering
  \includegraphics[width=0.98\textwidth]{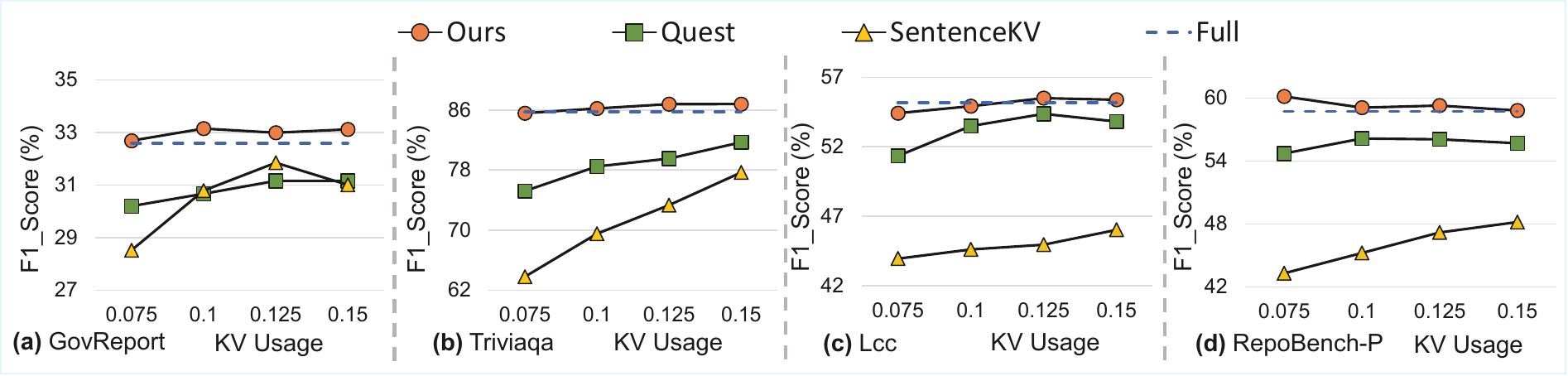}
  \setlength{\abovecaptionskip}{4pt} 
  \caption{Accuracy trends on LongBench under different KVCache usage.}
  \label{fig: Longbenc_usage}
\end{figure*}

As shown in the figure~\ref{fig: Longbenc_usage}, it illustrates how the accuracy of different methods changes with varying KVCache usage. DynSplit-KV demonstrates its advantage across different KVCache usage levels, and on some datasets, it achieves full-attention accuracy with less than 10\% KVCache usage.

\begin{figure}[!t]
  \centering
  \includegraphics[width=0.49\textwidth]{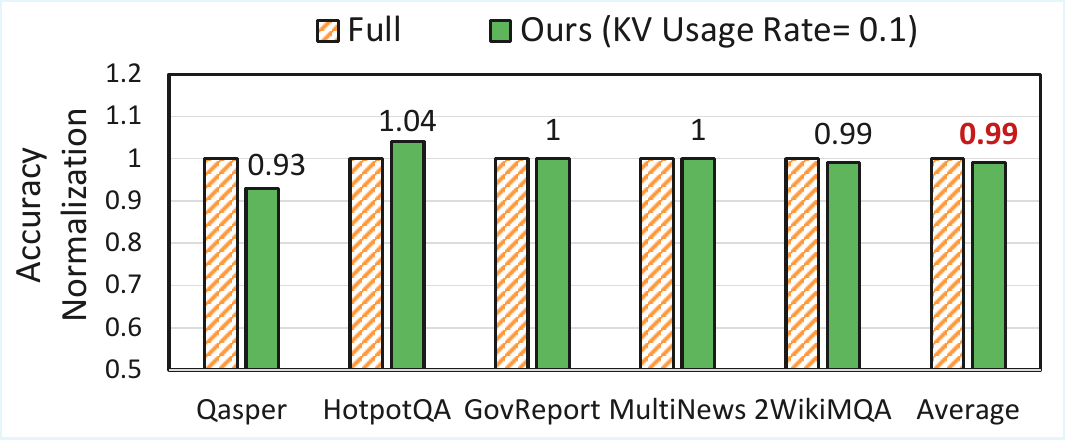}
  \vspace{-10pt}
  \setlength{\abovecaptionskip}{4pt} 
  \caption{Support for advanced GQA model (e.g., DeepSeek-R1-Distill-Llama-8B).}
  \vspace{-5pt}
  \label{fig: advanced}
\end{figure}

As shown in the Figure~\ref{fig: advanced}, we also evaluated the speed of our method on advanced GQA models. Taking DeepSeek-R1-Distill-Llama-8B as an example, at 10\% KVCache usage, our method achieves an average accuracy of 99\% of full attention performance across multiple LongBench datasets.

\textbf{LongBench V2.} We also evaluated the performance of DynSplit in ultra-long context inference scenarios ($> 64K$). As shown in the figure~\ref{tab:lbv2-half}, ours is able to maintain inference accuracy that is nearly comparable to the full KV cache baseline across different context length scenarios, while significantly reducing KVCache usage.

\begin{table}[H]
\small
\centering
\setlength{\tabcolsep}{6pt}
\caption{Llama-3-8B-1M  on LongBench v2.}
\begin{tabular}{@{}lccc@{}}
\toprule
\textbf{LongBench v2} & \multicolumn{2}{c}{\textbf{Accuracy normalization}} & \textbf{Avg. length} \\
\cmidrule(lr){2-3}
Split & Ours & Full & (tokens) \\
\midrule
Short   & 0.98 & 1.00 & 29,633 \\
Medium  & 0.97 & 1.00 & 93,909 \\
Overall & 0.98 & 1.00 & 56,882 \\
\bottomrule
\end{tabular}
\label{tab:lbv2-half}
\end{table}

\textbf{Passkey.} Since language modeling mainly relies on local dependencies, focusing on recent tokens suffices for good performance. However, long-range dependencies are crucial for long-text reasoning. Token-level KVCache eviction methods like StreamingLLM may discard KVCache needed for distant tokens, while block-level methods like Quest can lose semantic structure. To assess DynSplit-KV, we evaluate the passkey retrieval task, where models must find a passkey in large meaningless text. Answers are placed at varying depths, and models are tested under different KVCache budgets. We test Mistral-7B-Instruct-v0.2 on 10k/32k tokens and Llama-3-8B-1M on 100k tokens. Results in Table~\ref{tab: passkey} show DynSplit-KV achieves perfect accuracy with minimal budget (0.2\%). StreamingLLM fails if the passkey is outside the recent window, and Quest needs a larger budget.

\begin{table}[t]
\caption{Performance comparison across different sequence length budgets.}
\label{tab: passkey}
\centering
\scriptsize
\vspace{0.5em}
\begin{tabular}{@{}lccccc@{}}
\multicolumn{6}{c}{\textbf{Context length: 10k  //  Model: Mistral-7B-Instruct-v0.2}} \\
\toprule
Method / Budget & 36 & 64 & 128 & 256 & 512 \\
\midrule
StreamingLLM & 1\% & 1\% & 1\% & 3\% & 5\% \\
Quest & 43\% & 52\% & 92\% & 100\% & 100\% \\
Ours & \textbf{79\%} & \textbf{100\%} & \textbf{100\%} & \textbf{100\%} & \textbf{100\%} \\
\bottomrule
\end{tabular}
\vspace{1em}
\begin{tabular}{@{}lccccc@{}}
\multicolumn{6}{c}{\textbf{Context length: 30k  //  Model: Mistral-7B-Instruct-v0.2}} \\
\toprule
Method / Budget & 36 & 64 & 128 & 256 & 512 \\
\midrule
StreamingLLM & 1\% & 1\% & 1\% & 2\% & 4\% \\
Quest & 11\% & 15\% & 71\% & 100\% & 100\% \\
Ours & \textbf{44\%} & \textbf{100\%} & \textbf{100\%} & \textbf{100\%} & \textbf{100\%} \\
\bottomrule
\end{tabular}
\vspace{1em}
\begin{tabular}{@{}lccccc@{}}
\multicolumn{6}{c}{\textbf{Context length: 100k  //  Model: Llama-3-8B-1M}} \\
\toprule
Method / Budget & 256 & 512 & 1024 & 2048 & 4096 \\
\midrule
StreamingLLM & 1\% & 1\% & 1\% & 2\% & 4\% \\
Quest & 98\% & 100\% & 100\% & 100\% & 100\% \\
Ours & \textbf{100\%} & \textbf{100\%} & \textbf{100\%} & \textbf{100\%} & \textbf{100\%} \\
\bottomrule
\end{tabular}
\vspace{-0.5em}
\label{tab:performance_budget}
\end{table}

\subsection{Efficiency Evaluation}
\label{sec: efficiency evaluation}

\textbf{GPU Inference analysis.} To validate DynSplit-KV's practical acceleration, we compare it against a FlashAttention-based full KV cache baseline at 32K input length with output lengths ranging from 256 to 4096. By combining DynSplit-KV's efficient KV cache management with FlashAttention~\citep{dao2022flashattention}, we find that DynSplit-KV achieves significantly lower latency across all output lengths (Figure~8); while both methods' latency increases with longer outputs, DynSplit-KV's latency grows more slowly, attaining a maximum 2.16$\times$ ($\approx$ 2.2$\times$) inference speedup at 4096 output length.

\begin{figure}[H]
    \centering
    \includegraphics[width=\linewidth]{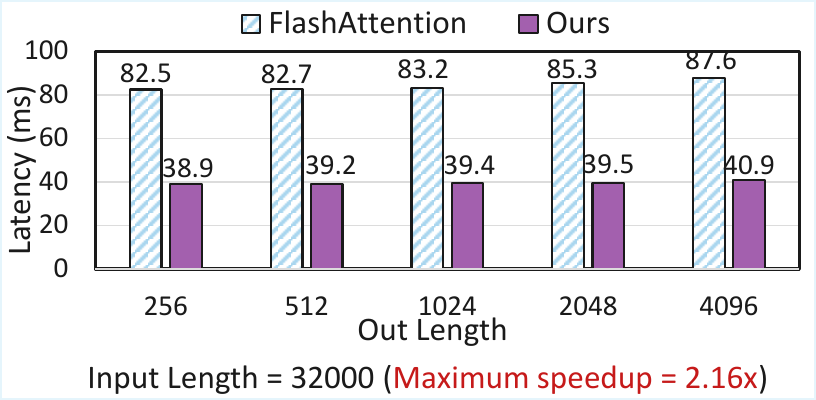}
    \caption{GPU Inference Efficiency Analysis.}
    \label{fig:CPU-GPU}
\end{figure}
 
\begin{figure}[H]
    \centering
    \includegraphics[width=\linewidth]{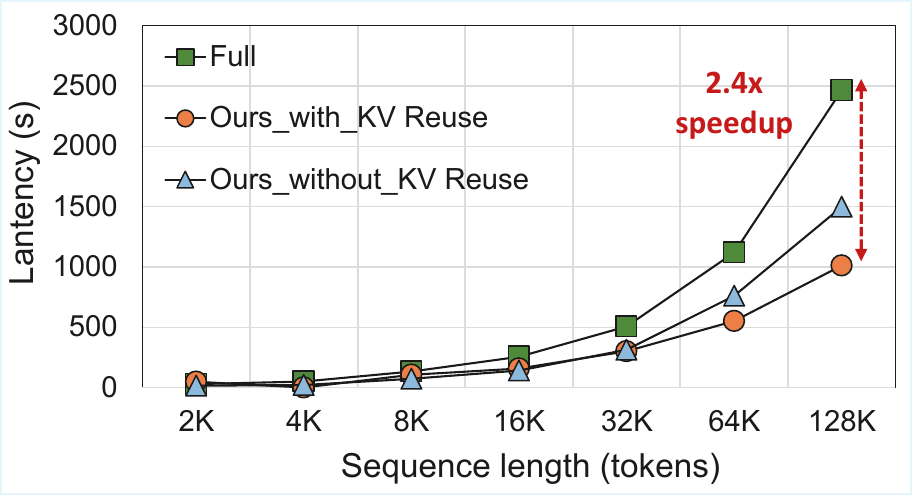}
    \vspace{-5pt}
    \caption{CPU-GPU Inference Efficiency Analysis.}
    \label{fig: CPU-GPU}
\end{figure}

\textbf{CPU-GPU Inference Analysis.} We implement KV offloading for our method to validate its acceleration in ultra-long context scenarios, where KVCache expands continuously and exceeds GPU memory with increasing context length. Experimental results in Figure~\ref{fig: CPU-GPU} show our method achieves a 2.4$\times$ maximum speedup over the full attention baseline in CPU-GPU deployment, with comparable accuracy.

\begin{figure}[H]
  \centering
  \includegraphics[width=0.49\textwidth]{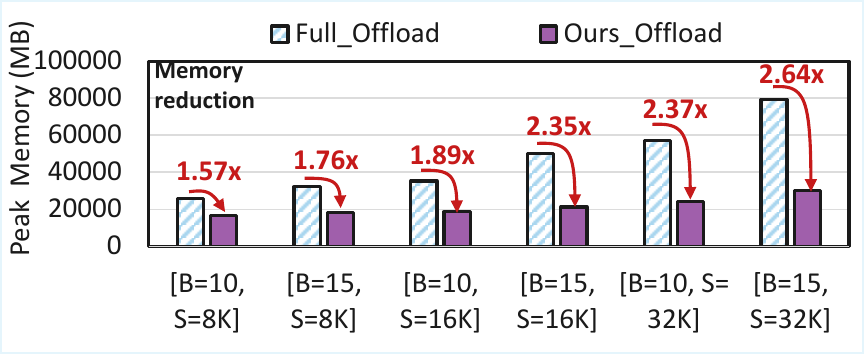}
  \vspace{-10pt}
  \setlength{\abovecaptionskip}{4pt} 
  \caption{Comparison of peak GPU memory between ours and full attention under different batches (B) and sequence lengths (S).}
  \vspace{-10pt}
  \label{fig: peak memory}
\end{figure}

\textbf{Peak Memory.} As shown in the Figure~\ref{fig: peak memory}, the peak memory savings of our method consistently increase with sequence length and batch size, exhibiting a scaling trend. In the case of a batch size of 15 and a sequence length of 32K, DynSplit-KV reduces peak memory by 2.64$\times$ ($\approx$ 2.6$\times$).

\subsection{Ablation Results}
\label{sec: ablation}

\begin{figure}[!t]
  \centering
  \includegraphics[width=0.49\textwidth]{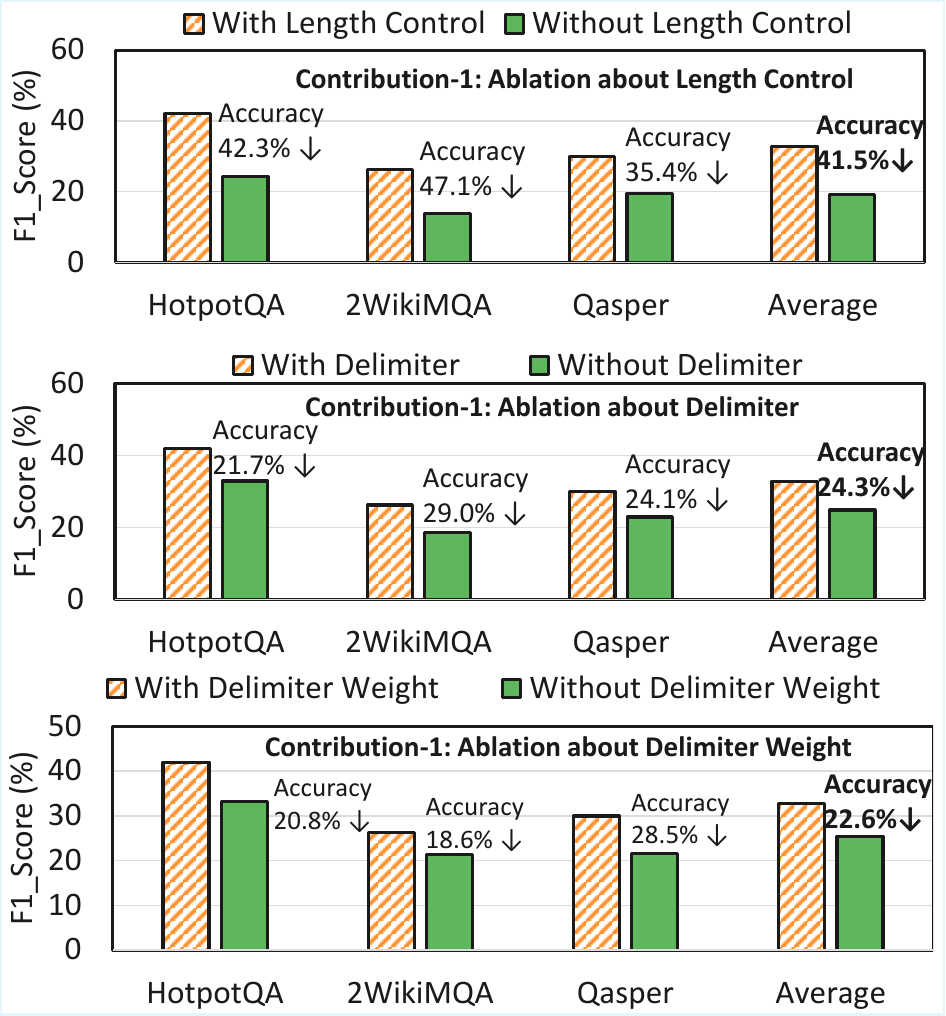}
  \vspace{-10pt}
  \setlength{\abovecaptionskip}{4pt} 
   \caption{Ablation study on method~\ref{sec: method2}, showing the effect of length control, delimiters, or delimiter selection on accuracy.}
  \vskip -0.2in
  \label{fig: ablation1}
\end{figure}

\textbf{Ablation1.} For Section~\ref{sec: method1}, two factors influence the dynamic calculation of delimiter importance: length and delimiter weight. This ablation experiment aims to demonstrate that dynamic KVCache splitting must be based on three types of information: length control, delimiters, and delimiter weights. As shown in Figure~\ref{fig: ablation1}, we conducted ablation experiments from three perspectives: \textit{without length control}, \textit{without delimiter}, and \textit{without delimiter weight}. The results show that the absence of any one of these three components leads to an average accuracy drop of more than 22\%. This indicates that the information to be considered for dynamic semantic splitting should not be limited to merely delimiter-based splitting or length control. 

\begin{figure}[!t]
  \centering
  \includegraphics[width=0.49\textwidth]{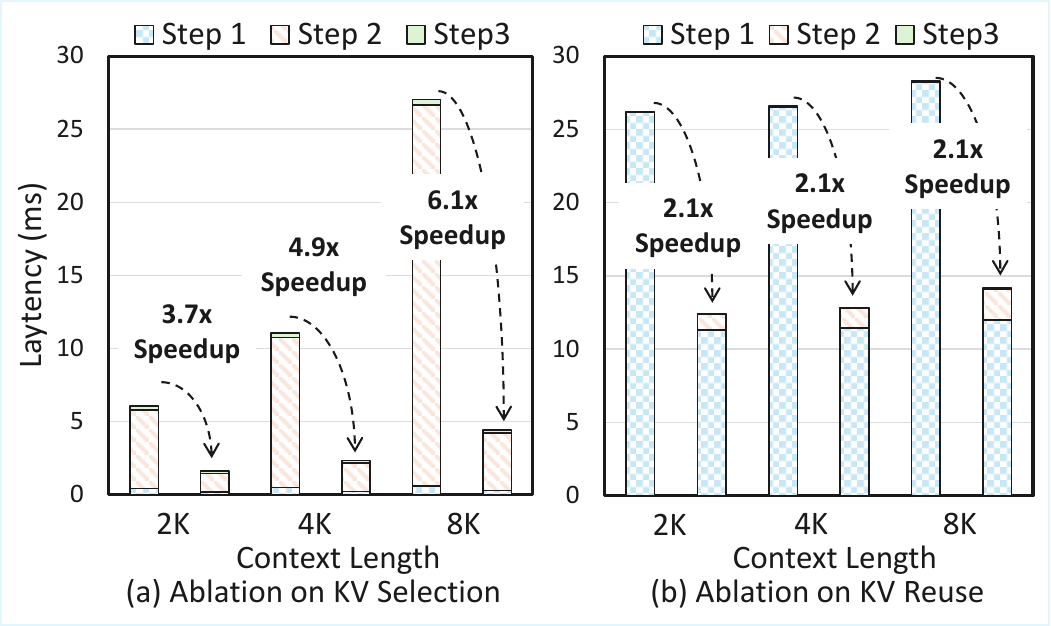}
  \vspace{-10pt}
  \setlength{\abovecaptionskip}{4pt} 
  \caption{Effect of variable-length processing on KVCache selection and reuse speedup.}
  \vspace{-10pt}
  \label{fig: ablation2}
\end{figure}

\textbf{Ablation2.} For Section~\ref{sec: method2}, we integrates KV Cache Selection and KV Cache Reuse techniques, so we conduct ablation experiments on both aspects to validate the effectiveness. As shown in Figure~\ref{fig: ablation2}, the design of a variable-length processing strategy is crucial for important KV selection and KV reuse under variable-length blocks situation. We conducted ablation experiments from two aspects: important KV selection and KV reuse. Experiments show that after adopting the variable-length processing strategy designed by us, the speedup ranges from 2.1x to 6.1x, which plays a significant role in long-context inference. The specific steps for Step 1, Step 2, and Step 3 are provided in the Appendix~\ref{B_2}. The additional overhead introduced here stems from Step 1. However, the experimental results show that while this additional overhead accounts for a very small proportion, it achieves significant overall acceleration.

\section{Conclusion}

We propose \textit{\textbf{DynSplit-KV}}, a KVCache compression method that dynamically identifies delimiters for semantic-aware splitting. We observe that rigid splitting leads to significant accuracy degradation across different scenarios, with relative drops ranging from 5.5\% to 55.1\%, due to the scenario-dependent nature of semantic boundaries. This observation highlights the necessity of \textbf{dynamic semantic splitting} to better align with semantics. However, we still face two main challenges: improper delimiter selection misaligns semantics with the KVCache, and variable-length blocks after splitting introduce over 73.1\% additional inference overhead. To address these, we propose \textit{DynSplit-KV}, which dynamically identifies delimiters and maps variable-length blocks to a fixed-length representation. Experiments show that \textit{DynSplit-KV} achieves the highest accuracy, 2.2$\times$ speedup compared with FlashAttention and 2.6$\times$ peak memory reduction in long-context scenarios. We will continue to explore KVCache compression based on dynamic semantic splitting in future work.

\nocite{langley00}

\bibliography{main}

\begin{thebibliography}{29}
\providecommand{\natexlab}[1]{#1}
\providecommand{\url}[1]{\texttt{#1}}
\expandafter\ifx\csname urlstyle\endcsname\relax
  \providecommand{\doi}[1]{doi: #1}\else
  \providecommand{\doi}{doi: \begingroup \urlstyle{rm}\Url}\fi

\bibitem[Anthropic(2025)]{anthropic2025claude37}
Anthropic.
\newblock Claude 3.7 sonnet and claude code, February 2025.
\newblock URL \url{https://www.anthropic.com/news/claude-3-7-sonnet}.
\newblock Online; accessed 25-September-2025.

\bibitem[{Anthropic}(2025)]{anthropic2025research}
{Anthropic}.
\newblock Claude takes research to new places, 2025.
\newblock URL \url{https://www.anthropic.com/news/research}.
\newblock Online; accessed 25-September-2025.

\bibitem[{Anthropic}({2025})]{claudecoder2025}
{Anthropic}.
\newblock {Claude Coder}, {2025}.
\newblock URL \url{https://claude.com/product/claude-code}.
\newblock Online; accessed 25-September-2025.

\bibitem[Bai et~al.(2023)Bai, Lv, Zhang, Lyu, Tang, Huang, Du, Liu, Zeng, Hou, et~al.]{bai2023longbench}
Bai, Y., Lv, X., Zhang, J., Lyu, H., Tang, J., Huang, Z., Du, Z., Liu, X., Zeng, A., Hou, L., et~al.
\newblock Longbench: A bilingual, multitask benchmark for long context understanding.
\newblock \emph{arXiv preprint arXiv:2308.14508}, 2023.

\bibitem[Bai et~al.(2025)Bai, Tu, Zhang, Peng, Wang, Lv, Cao, Xu, Hou, Dong, et~al.]{bai2025longbench}
Bai, Y., Tu, S., Zhang, J., Peng, H., Wang, X., Lv, X., Cao, S., Xu, J., Hou, L., Dong, Y., et~al.
\newblock Longbench v2: Towards deeper understanding and reasoning on realistic long-context multitasks.
\newblock In \emph{Proceedings of the 63rd Annual Meeting of the Association for Computational Linguistics (Volume 1: Long Papers)}, pp.\  3639--3664, 2025.

\bibitem[Chen et~al.(2025)Chen, Liu, Xu, Gao, and Wang]{chen2025sablock}
Chen, J., Liu, J., Xu, H., Gao, X., and Wang, S.
\newblock Sablock: Semantic-aware kv cache eviction with adaptive compression block size.
\newblock \emph{arXiv preprint arXiv:2510.22556}, 2025.

\bibitem[Comanici et~al.(2025)Comanici, Bieber, Schaekermann, Pasupat, Sachdeva, Dhillon, Blistein, Ram, Zhang, Rosen, et~al.]{comanici2025gemini}
Comanici, G., Bieber, E., Schaekermann, M., Pasupat, I., Sachdeva, N., Dhillon, I., Blistein, M., Ram, O., Zhang, D., Rosen, E., et~al.
\newblock Gemini 2.5: Pushing the frontier with advanced reasoning, multimodality, long context, and next generation agentic capabilities.
\newblock \emph{arXiv preprint arXiv:2507.06261}, 2025.

\bibitem[Dao et~al.(2022)Dao, Fu, Ermon, Rudra, and R{\'e}]{dao2022flashattention}
Dao, T., Fu, D., Ermon, S., Rudra, A., and R{\'e}, C.
\newblock Flashattention: Fast and memory-efficient exact attention with io-awareness.
\newblock \emph{Advances in neural information processing systems}, 35:\penalty0 16344--16359, 2022.

\bibitem[Guo et~al.(2025)Guo, Yang, Zhang, Song, Wang, Zhu, Xu, Zhang, Ma, Bi, Zhang, Yu, Wu, Wu, Gou, Shao, Li, Gao, Liu, Xue, Wang, Wu, Feng, Lu, Zhao, Deng, Ruan, Dai, Chen, Ji, Li, Lin, Dai, Luo, Hao, Chen, Li, Zhang, Xu, Ding, Gao, Qu, Li, Guo, Li, Chen, Yuan, Tu, Qiu, Li, Cai, Ni, Liang, Chen, Dong, Hu, You, Gao, Guan, Huang, Yu, Wang, Zhang, Zhao, Wang, Zhang, Xu, Xia, Zhang, Zhang, Tang, Zhou, Li, Wang, Li, Tian, Huang, Zhang, Wang, Chen, Du, Ge, Zhang, Pan, Wang, Chen, Jin, Chen, Lu, Zhou, Chen, Ye, Wang, Yu, Zhou, Pan, Li, Zhou, Wu, Yun, Pei, Sun, Wang, Zeng, Liu, Liang, Gao, Yu, Zhang, Xiao, An, Liu, Wang, Chen, Nie, Cheng, Liu, Xie, Liu, Yang, Li, Su, Lin, Li, Jin, Shen, Chen, Sun, Wang, Song, Zhou, Wang, Shan, Li, Wang, Wei, Zhang, Xu, Li, Zhao, Sun, Wang, Yu, Zhang, Shi, Xiong, He, Piao, Wang, Tan, Ma, Liu, Guo, Ou, Wang, Gong, Zou, He, Xiong, Luo, You, Liu, Zhou, Zhu, Huang, Li, Zheng, Zhu, Ma, Tang, Zha, Yan, Ren, Ren, Sha, Fu, Xu, Xie, Zhang, Hao, Ma, Yan, Wu, Gu, Zhu, Liu, Li, Xie, Song,
  Pan, Huang, Xu, Zhang, and Zhang]{Guo_2025}
Guo, D., Yang, D., Zhang, H., Song, J., Wang, P., Zhu, Q., Xu, R., Zhang, R., Ma, S., Bi, X., Zhang, X., Yu, X., Wu, Y., Wu, Z.~F., Gou, Z., Shao, Z., Li, Z., Gao, Z., Liu, A., Xue, B., Wang, B., Wu, B., Feng, B., Lu, C., Zhao, C., Deng, C., Ruan, C., Dai, D., Chen, D., Ji, D., Li, E., Lin, F., Dai, F., Luo, F., Hao, G., Chen, G., Li, G., Zhang, H., Xu, H., Ding, H., Gao, H., Qu, H., Li, H., Guo, J., Li, J., Chen, J., Yuan, J., Tu, J., Qiu, J., Li, J., Cai, J.~L., Ni, J., Liang, J., Chen, J., Dong, K., Hu, K., You, K., Gao, K., Guan, K., Huang, K., Yu, K., Wang, L., Zhang, L., Zhao, L., Wang, L., Zhang, L., Xu, L., Xia, L., Zhang, M., Zhang, M., Tang, M., Zhou, M., Li, M., Wang, M., Li, M., Tian, N., Huang, P., Zhang, P., Wang, Q., Chen, Q., Du, Q., Ge, R., Zhang, R., Pan, R., Wang, R., Chen, R.~J., Jin, R.~L., Chen, R., Lu, S., Zhou, S., Chen, S., Ye, S., Wang, S., Yu, S., Zhou, S., Pan, S., Li, S.~S., Zhou, S., Wu, S., Yun, T., Pei, T., Sun, T., Wang, T., Zeng, W., Liu, W., Liang, W., Gao, W., Yu, W.,
  Zhang, W., Xiao, W.~L., An, W., Liu, X., Wang, X., Chen, X., Nie, X., Cheng, X., Liu, X., Xie, X., Liu, X., Yang, X., Li, X., Su, X., Lin, X., Li, X.~Q., Jin, X., Shen, X., Chen, X., Sun, X., Wang, X., Song, X., Zhou, X., Wang, X., Shan, X., Li, Y.~K., Wang, Y.~Q., Wei, Y.~X., Zhang, Y., Xu, Y., Li, Y., Zhao, Y., Sun, Y., Wang, Y., Yu, Y., Zhang, Y., Shi, Y., Xiong, Y., He, Y., Piao, Y., Wang, Y., Tan, Y., Ma, Y., Liu, Y., Guo, Y., Ou, Y., Wang, Y., Gong, Y., Zou, Y., He, Y., Xiong, Y., Luo, Y., You, Y., Liu, Y., Zhou, Y., Zhu, Y.~X., Huang, Y., Li, Y., Zheng, Y., Zhu, Y., Ma, Y., Tang, Y., Zha, Y., Yan, Y., Ren, Z.~Z., Ren, Z., Sha, Z., Fu, Z., Xu, Z., Xie, Z., Zhang, Z., Hao, Z., Ma, Z., Yan, Z., Wu, Z., Gu, Z., Zhu, Z., Liu, Z., Li, Z., Xie, Z., Song, Z., Pan, Z., Huang, Z., Xu, Z., Zhang, Z., and Zhang, Z.
\newblock Deepseek-r1 incentivizes reasoning in llms through reinforcement learning.
\newblock \emph{Nature}, 645\penalty0 (8081):\penalty0 633–638, September 2025.
\newblock ISSN 1476-4687.
\newblock \doi{10.1038/s41586-025-09422-z}.
\newblock URL \url{http://dx.doi.org/10.1038/s41586-025-09422-z}.

\bibitem[Hooper et~al.(2024)Hooper, Kim, Mohammadzadeh, Mahoney, Shao, Keutzer, and Gholami]{hooper2024kvquant}
Hooper, C., Kim, S., Mohammadzadeh, H., Mahoney, M.~W., Shao, Y.~S., Keutzer, K., and Gholami, A.
\newblock Kvquant: Towards 10 million context length llm inference with kv cache quantization.
\newblock \emph{Advances in Neural Information Processing Systems}, 37:\penalty0 1270--1303, 2024.

\bibitem[Jiang et~al.(2023)Jiang, Sablayrolles, Mensch, Bamford, Chaplot, de~las Casas, Bressand, Lengyel, Lample, Saulnier, Lavaud, Lachaux, Stock, Scao, Lavril, Wang, Lacroix, and Sayed]{jiang2023mistral7b}
Jiang, A.~Q., Sablayrolles, A., Mensch, A., Bamford, C., Chaplot, D.~S., de~las Casas, D., Bressand, F., Lengyel, G., Lample, G., Saulnier, L., Lavaud, L.~R., Lachaux, M.-A., Stock, P., Scao, T.~L., Lavril, T., Wang, T., Lacroix, T., and Sayed, W.~E.
\newblock Mistral 7b, 2023.
\newblock URL \url{https://arxiv.org/abs/2310.06825}.

\bibitem[Langley(2000)]{langley00}
Langley, P.
\newblock Crafting papers on machine learning.
\newblock In Langley, P. (ed.), \emph{Proceedings of the 17th International Conference on Machine Learning (ICML 2000)}, pp.\  1207--1216, Stanford, CA, 2000. Morgan Kaufmann.

\bibitem[Li et~al.(2023)Li, Shao, Xie, Sheng, Zheng, Gonzalez, Stoica, Ma, and Zhang]{li2023long}
Li, D., Shao, R., Xie, A., Sheng, Y., Zheng, L., Gonzalez, J., Stoica, I., Ma, X., and Zhang, H.
\newblock How long can context length of open-source llms truly promise?
\newblock In \emph{NeurIPS 2023 Workshop on Instruction Tuning and Instruction Following}, 2023.

\bibitem[Li et~al.(2024)Li, Huang, Yang, Venkitesh, Locatelli, Ye, Cai, Lewis, and Chen]{li2024snapkv}
Li, Y., Huang, Y., Yang, B., Venkitesh, B., Locatelli, A., Ye, H., Cai, T., Lewis, P., and Chen, D.
\newblock Snapkv: Llm knows what you are looking for before generation.
\newblock \emph{Advances in Neural Information Processing Systems}, 37:\penalty0 22947--22970, 2024.

\bibitem[Liu et~al.(2024{\natexlab{a}})Liu, Li, Zhao, Zhang, and Guo]{liu2024clusterkv}
Liu, G., Li, C., Zhao, J., Zhang, C., and Guo, M.
\newblock Clusterkv: Manipulating llm kv cache in semantic space for recallable compression.
\newblock \emph{arXiv preprint arXiv:2412.03213}, 2024{\natexlab{a}}.

\bibitem[Liu et~al.(2025)Liu, Tang, Dong, Li, Liu, Li, Hu, and Chu]{liu2025chunkkv}
Liu, X., Tang, Z., Dong, P., Li, Z., Liu, Y., Li, B., Hu, X., and Chu, X.
\newblock Chunkkv: Semantic-preserving kv cache compression for efficient long-context llm inference.
\newblock \emph{arXiv preprint arXiv:2502.00299}, 2025.

\bibitem[Liu et~al.(2024{\natexlab{b}})Liu, Yuan, Jin, Zhong, Xu, Braverman, Chen, and Hu]{liu2024kivi}
Liu, Z., Yuan, J., Jin, H., Zhong, S., Xu, Z., Braverman, V., Chen, B., and Hu, X.
\newblock Kivi: a tuning-free asymmetric 2bit quantization for kv cache.
\newblock In \emph{Proceedings of the 41st International Conference on Machine Learning}, pp.\  32332--32344, 2024{\natexlab{b}}.

\bibitem[{OpenAI}(2025)]{openai2025deepresearch}
{OpenAI}.
\newblock Introducing deep research, January 2025.
\newblock URL \url{https://openai.com/index/introducing-deep-research/}.
\newblock Online; accessed 25-September-2025.

\bibitem[Pekelis et~al.(2024)Pekelis, Feil, Moret, Huang, and Peng]{gradientlongcontextllama3}
Pekelis, L., Feil, M., Moret, F., Huang, M., and Peng, T.
\newblock Llama 3 gradient: A series of long context models, 2024.
\newblock URL \url{https://gradient.ai/blog/scaling-rotational-embeddings-for-long-context-language-models}.

\bibitem[Peng et~al.(2023)Peng, Quesnelle, Fan, and Shippole]{peng2023yarn}
Peng, B., Quesnelle, J., Fan, H., and Shippole, E.
\newblock Yarn: Efficient context window extension of large language models.
\newblock \emph{arXiv preprint arXiv:2309.00071}, 2023.

\bibitem[{Qwen Team}(2025)]{qwen3coder2025}
{Qwen Team}.
\newblock {Qwen3-Coder: Agentic Coding in the World}, 2025.
\newblock URL \url{https://qwen.ai/blog?id=d927d7d2e59d059045ce758ded34f98c0186d2d7&from=research.research-list}.
\newblock Online; accessed 25-September-2025.

\bibitem[Sun et~al.(2024)Sun, Chang, Bao, Zheng, Zheng, Liu, Dong, Chi, and Chen]{sun2024shadowkv}
Sun, H., Chang, L.-W., Bao, W., Zheng, S., Zheng, N., Liu, X., Dong, H., Chi, Y., and Chen, B.
\newblock Shadowkv: Kv cache in shadows for high-throughput long-context llm inference.
\newblock \emph{arXiv preprint arXiv:2410.21465}, 2024.

\bibitem[Tang et~al.(2024)Tang, Zhao, Zhu, Xiao, Kasikci, and Han]{tang2024quest}
Tang, J., Zhao, Y., Zhu, K., Xiao, G., Kasikci, B., and Han, S.
\newblock Quest: Query-aware sparsity for efficient long-context llm inference.
\newblock \emph{arXiv preprint arXiv:2406.10774}, 2024.

\bibitem[Xiao et~al.(2024)Xiao, Zhang, Han, Xiao, Lin, Zhang, Liu, and Sun]{xiao2024infllm}
Xiao, C., Zhang, P., Han, X., Xiao, G., Lin, Y., Zhang, Z., Liu, Z., and Sun, M.
\newblock Infllm: Training-free long-context extrapolation for llms with an efficient context memory.
\newblock \emph{Advances in Neural Information Processing Systems}, 37:\penalty0 119638--119661, 2024.

\bibitem[Xiao et~al.(2023{\natexlab{a}})Xiao, Lin, Seznec, Wu, Demouth, and Han]{xiao2023smoothquant}
Xiao, G., Lin, J., Seznec, M., Wu, H., Demouth, J., and Han, S.
\newblock Smoothquant: Accurate and efficient post-training quantization for large language models.
\newblock In \emph{International conference on machine learning}, pp.\  38087--38099. PMLR, 2023{\natexlab{a}}.

\bibitem[Xiao et~al.(2023{\natexlab{b}})Xiao, Tian, Chen, Han, and Lewis]{xiao2023efficient}
Xiao, G., Tian, Y., Chen, B., Han, S., and Lewis, M.
\newblock Efficient streaming language models with attention sinks.
\newblock \emph{arXiv preprint arXiv:2309.17453}, 2023{\natexlab{b}}.

\bibitem[Zhang et~al.(2024)Zhang, Ji, Chen, Fu, Miao, Nie, Chen, and Cui]{zhang2024pqcache}
Zhang, H., Ji, X., Chen, Y., Fu, F., Miao, X., Nie, X., Chen, W., and Cui, B.
\newblock Pqcache: Product quantization-based kvcache for long context llm inference.
\newblock \emph{arXiv preprint arXiv:2407.12820}, 2024.

\bibitem[Zhang et~al.(2023)Zhang, Sheng, Zhou, Chen, Zheng, Cai, Song, Tian, R{\'e}, Barrett, et~al.]{zhang2023h2o}
Zhang, Z., Sheng, Y., Zhou, T., Chen, T., Zheng, L., Cai, R., Song, Z., Tian, Y., R{\'e}, C., Barrett, C., et~al.
\newblock H2o: Heavy-hitter oracle for efficient generative inference of large language models.
\newblock \emph{Advances in Neural Information Processing Systems}, 36:\penalty0 34661--34710, 2023.

\bibitem[Zhu et~al.(2025)Zhu, Falahati, Yang, and Amiri]{zhu2025sentencekv}
Zhu, Y., Falahati, A., Yang, D.~H., and Amiri, M.~M.
\newblock Sentencekv: Efficient llm inference via sentence-level semantic kv caching.
\newblock \emph{arXiv preprint arXiv:2504.00970}, 2025.

\end{thebibliography}
\bibliographystyle{icml2026}

\newpage
\appendix
\onecolumn

\begin{center}
    \huge\bfseries Appendix 
\end{center}

Outline: In Appendix~\ref{Appendix A}, we primarily provide additional experiments, including: (1) extra profiling of different stages during inference, showing the proportion of attention time and KVCache memory movement time in large language model inference, to further highlight the importance of loading fewer KVCache entries during the attention stage for accelerating LLM inference in long-context scenarios, (2) additional ablation studies: our main result uses element-wise min/max for KVCache compression and (3) Additional main result for accuracy under different KVCache usage. To demonstrate the generality of our segmentation approach across different compression methods, we also evaluate average pooling compression. In Appendix~\ref{Appendix B}, we mainly provide implementation details.

\section{Additional Experiments.}
\label{Appendix A}

\subsection{Time profiling for long-context inference.}
\label{A_1}

In LLM inference, Transformer layers are composed of Attention (Attn.) and Feed-Forward Network (FFN) modules. Attention computes token-wise interactions by matching queries with keys to weight values, capturing contextual dependencies, while FFN applies nonlinear transformations independently to each token. During autoregressive inference, the KVCache stores the keys and values of previously generated tokens to avoid recomputation in Attention. However, the time required to load KVCache into memory for each step can significantly impact the overall latency of the Attention phase, especially in long-context scenarios. Table~\ref{tab:sdpa_overhead_A800} illustrates the time breakdown between the attention module and the FFN during inference.

For Table~\ref{tab:sdpa_overhead_A800}, experiments were conducted using PyTorch to run the Llama3-8B model. The results indicate that for sequence lengths exceeding 1K tokens, attention accounts for the majority of inference time, exceeding 3.5$\times$ the proportion of FFN computation. This highlights the importance of optimizing attention for long-context inference.

\begin{table}[h] 
\centering
\caption{Time Cost Ratios of SDPA Attention on Nvidia A800 (SDPA stands for Scaled Dot-Product Attention, a commonly used attention mechanism in Transformers.)}
\label{tab:sdpa_overhead_A800}
\begin{tabular}{ccccc} 
\toprule
Context & Device      & Attn. Type & Attn./FFN & FFN/Total  \\ 
\hline\hline
1k      & Nvidia A800 & sdpa       & 3.96      & 5.00\%     \\ 
\hline\hline
2k      & Nvidia A800 & sdpa       & 3.86      & 4.00\%     \\ 
\hline\hline
3k      & Nvidia A800 & sdpa       & 3.62      & 3.78\%     \\ 
\hline\hline
4k      & Nvidia A800 & sdpa       & 3.55      & 3.30\%     \\ 
\hline\hline
5k      & Nvidia A800 & sdpa       & 4.8       & 2.74\%     \\ 
\hline\hline
6k      & Nvidia A800 & sdpa       & 4.74      & 2.61\%     \\ 
\hline\hline
7k      & Nvidia A800 & sdpa       & 4.74      & 2.42\%     \\ 
\hline\hline
8k      & Nvidia A800 & sdpa       & 4.82      & 2.32\%     \\ 
\hline\hline
9k      & Nvidia A800 & sdpa       & 4.86      & 2.22\%     \\ 
\hline\hline
10k     & Nvidia A800 & sdpa       & 4.8       & 2.00\%     \\
\bottomrule
\end{tabular}
\end{table}


The memory footprint of the KVCache grows proportionally with the sequence length. When the KV cache exceeds a certain length, it no longer fits in GPU memory, necessitating its placement on the CPU and the use of CPU–GPU collaborative inference. Table~\ref{tab:kv_transfer_gpu_computation} reports the KVCache transfer time in long-context CPU–GPU collaborative inference scenarios.

For Table~\ref{tab:kv_transfer_gpu_computation}, the experimental setup uses Llama2-13B with a hidden size of 5120 and 40 attention heads. The KVCache is fully placed on the CPU, and the attention time for a single decoding step is measured using an NVIDIA A800 GPU with FP32 inference. The experimental results show that in the CPU–GPU collaborative setting, due to the limited PCIe bandwidth between the CPU and GPU, the overhead of transferring the required KV cache from the CPU (storage unit) to the GPU (computation unit) during attention computation becomes the primary bottleneck, accounting for over 99\% of the total time at a context length of 128K.

Tables~\ref{tab:sdpa_overhead_A800} and Table~\ref{tab:kv_transfer_gpu_computation} collectively show that in long-context scenarios, the amount of KV cache transferred accounts for over 90\% of the factors affecting inference time, indicating that generating tokens with as little KV cache as possible can lead to direct and substantial speedups.


\begin{table}
\centering
\footnotesize
\caption{KV Transfer Time, GPU Computation Time, and Transfer Time Proportion Under Different Context Lengths}
\label{tab:kv_transfer_gpu_computation}
\begin{tabular}{cccc} 
\toprule
Context & KV transfer time [CPU-GPU] (ms) & GPU computation time~(ms) & Proportion of transfer time  \\ 
\hline\hline
50      & 0.69                            & 0.42                      & 0.621621622                  \\ 
\hline\hline
100     & 1.12                            & 0.2                       & 0.848484848                  \\ 
\hline\hline
200     & 2.27                            & 0.39                      & 0.853383459                  \\ 
\hline\hline
500     & 5.35                            & 0.38                      & 0.933682373                  \\ 
\hline\hline
1k      & 9.3                             & 0.64                      & 0.935613682                  \\ 
\hline\hline
2k      & 16.61                           & 0.67                      & 0.961226852                  \\ 
\hline\hline
3k      & 25.15                           & 0.76                      & 0.970667696                  \\ 
\hline\hline
4k      & 30.81                           & 0.81                      & 0.974383302                  \\ 
\hline\hline
5k      & 37.6                            & 0.7                       & 0.981979629                  \\ 
\hline\hline
6k      & 45.94                           & 0.86                      & 0.981623932                  \\ 
\hline\hline
7k      & 52.88                           & 0.9                       & 0.983265154                  \\ 
\hline\hline
8k      & 60.27                           & 0.82                      & 0.986577181                  \\ 
\hline\hline
9k      & 67.45                           & 0.93                      & 0.986399532                  \\ 
\hline\hline
10k     & 75.83                           & 0.85                      & 0.988914971                  \\ 
\hline\hline
16k     & 104.74                          & 0.86                      & 0.991856061                  \\ 
\hline\hline
32k     & 208.95                          & 1.16                      & 0.994526416                  \\ 
\hline\hline
50k     & 365.54                          & 1.35                      & 0.99634758                   \\ 
\hline\hline
64k     & 520.84                          & 1.94                      & 0.99628907                   \\ 
\hline\hline
128k    & 918.78                          & 2.16                      & 0.99765457                   \\
\bottomrule
\end{tabular}
\end{table}

\subsection{Additional ablation study}
\label{A_2}

To demonstrate that the observed accuracy improvements stem from our segmentation design rather than the compression strategy, we replace the original element-wise min/max KVCache compression with mean pooling, while keeping all other settings unchanged. The results are shown in the Table~\ref{tab: mean pooling}. The results show that the average accuracy remains nearly unchanged (43.64 v.s. 42.46) across different compression methods, indicating that our segmentation method is robust to the choice of compression strategy.

\begin{table*}[ht]
\centering
\scriptsize
\caption{Main accuracy results (KV Usage Rate = 0.1).}
\label{tab: mean pooling}
\begin{tabular}{c|c|cc|cc|cc|c|cc|>{\columncolor{gray!15}}c} 
\toprule[1.5pt]
& \multicolumn{1}{c|}{} 
  & \multicolumn{2}{c|}{Single-Document QA} 
  & \multicolumn{2}{c|}{Multi-Document QA} 
  & \multicolumn{2}{c|}{Summarization} 
  & Few-shot & \multicolumn{2}{c|}{Code}  & \cellcolor{gray!15} \\
\cmidrule(lr){3-4} \cmidrule(lr){5-6} \cmidrule(lr){7-8} \cmidrule(lr){9-9} \cmidrule(lr){10-11}
\multicolumn{1}{c|}{\begin{tabular}{c}KV Usage\\Rate = 0.1\end{tabular}}
& Method 
& NarrativeQA & Qasper 
& HotpotQA & 2WikiMQA 
& Gov-R & MultiNews 
& TriviaQA & Lcc & R-P & \cellcolor{gray!15}\textbf{Avg.} \\ 
\midrule[1.2pt]

\rowcolor{cyan!15}
Ours & Max/Min    
& 25.47 & 30.20 
& 42.01 & 26.26 
& 33.15 & 26.40
& 86.20 & 54.92 & 59.11   
& \cellcolor{gray!15}42.64    \\ 

\midrule[0.8pt]
\rowcolor{cyan!15}
Ours & Mean pooling        
& 23.63 & 31.30 
& 40.22 & 26.65 
& 32.79 & 26.72 
& 85.94 & 56.14 & 58.75   
& \cellcolor{gray!15}42.46    \\

\midrule[0.8pt]

Full KV & Full         & 26.47 & 32.91 & 43.72 & 26.97 & 32.59 & 26.97 & 85.74 & 55.18 & 53.94   & \cellcolor{gray!15}42.72  \\

\bottomrule[1.5pt]
\end{tabular}
\end{table*}

\subsection{Additional main result}
\label{A_3}

\begin{figure*}[!t]  
  \centering
  \includegraphics[width=0.98\textwidth]{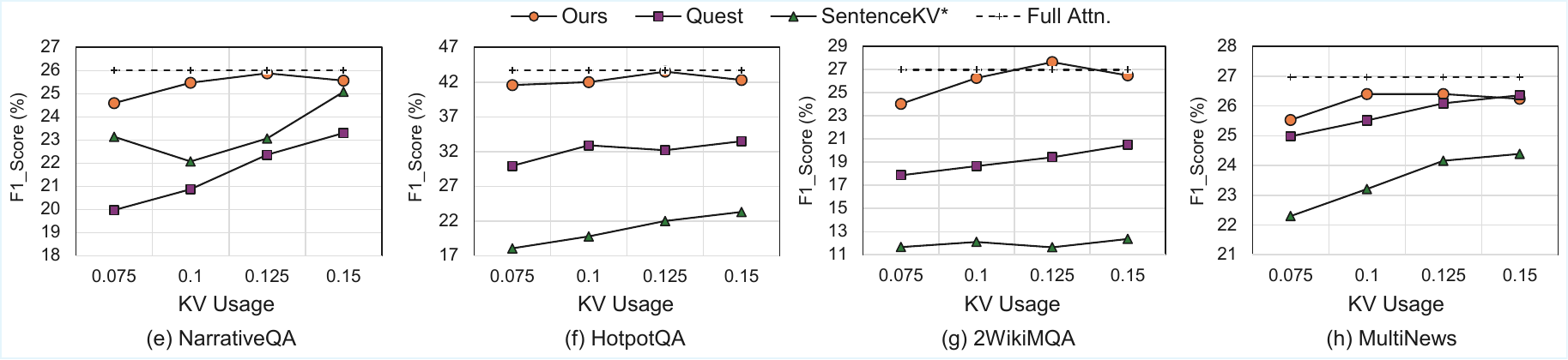}
  \vspace{5pt}
  \setlength{\abovecaptionskip}{4pt} 
  \caption{Accuracy trends on LongBench under different KVCache usage.}
  \label{fig: Longbenc_usage appendix}
  \vspace{5pt}
\end{figure*}

We not only evaluated the accuracy of our method at a KVCache usage rate of 0.1, but also tested its performance on LongBench across eight datasets with KVCache usage rates of 0.075, 0.1, 0.125, and 0.15. Four of these datasets are presented in Section~\ref{sec: experiment} of the main text, while the remaining four are shown in the Figure~\ref{fig: Longbenc_usage appendix}. The experimental results show that across nearly all KVCache usage rates, our method achieves the highest accuracy, and in some compression settings and datasets, it even surpasses full attention. This demonstrates the potential of our approach to effectively utilize KVCache compression in long-context inference scenarios.

\newpage
\section{Experimental Details}
\label{Appendix B}

\subsection{Delimiter Importance example}
\label{B_1}

As shown in the Table~\ref{tab: delimiter importance example}, we present example delimiter importance scores for Mistral-7B-Instruct-v0.2. The detailed computation procedure is described in Section~\ref{sec: observations}, Delimiter Importance Estimation via Attention Dependency.

\begin{table}[H]
\caption{Delimiters importance (For example, Mistral-7B-Instruct-v0.2).}
\label{tab: delimiter importance example}
\centering
\small
\setlength{\tabcolsep}{5pt}
\begin{tabular}{@{}lccccccccccccc@{}}
\toprule
\multicolumn{13}{c}{\textbf{Mistral-7B-Instruct-v0.2}} \\
\midrule
Delimiters   & . & ! & ? & \dots & ; & : & , & ' & " & ( & ) & [ & ] \\
\midrule
Token id & 28723 & 609 & 28804 & 1101 & 28745 & 28747 & 28725 & 28742 & 28808 & 28732 & 557 & 28792 & 28793 \\
Weight   & 1.0   & 1.0 & 0.9   & 1.0  & 0.7   & 0.7   & 0.6   & 0.5   & 0.9   & 0.5   & 0.6   & 0.5   & 0.5   \\
\bottomrule
\end{tabular}
\end{table}

\subsection{Implementation details}
\label{B_2}

In order to ensure reproducibility, we will release the code. The implementation details are as follows:

\textbf{FlashAttention combination.} We implement our method using PyTorch. To integrate with FlashAttention, we leverage its attention operator to accelerate the $Q \cdot K^\top$ matrix multiplication and Softmax computation. Specifically, FlashAttention is applied during the attention computation over the selected important KVCache.

\textbf{KVCache Selection with V2F (section~\ref{sec: method2}) proceeds as follows:}
\label{KV Selection with V2F}

\begin{enumerate}
    \item \textbf{Step1-Compute attention scores for each semantic chunk:} Based on the attention values between each semantic chunk and the query, we calculate the attention scores in the shape of \([B, H, Q=1, S]\). Then, we apply a top-k operation to obtain the top-k token indices (kvcache indices) in the shape of \([B, H, Q=1, \text{token\_budget}]\), where \(k\) is dynamically controlled by the token budget.
    
    \item \textbf{Step2-Parallel loading of required KV cache:} We then parallelly load only the required KV cache based on the top-k token indices.
    
    \item \textbf{Step3-Query @ required Key and calculate attention:} For the query and the required key, we compute the attention weights in the shape of \([B, H, Q=1, \text{token\_budget}]\), apply softmax to the attention weights, and finally perform the attention operation between the attention weights and the required values to obtain the output.
\end{enumerate}

\textbf{KVCache Reuse with V2F (section~\ref{sec: method2}) proceeds as follows:}
\label{KV Reuse with V2F}

\begin{enumerate}
    \item \textbf{Step1-Iterate through each attention head:} We compute the reusable KV for each head serially. Across multiple heads, we truncate the reusable KV based on the minimum reusable data volume. This ensures a consistent length of reusable KV caches among all heads, overcoming the inconsistency caused by variable-length blocks.
    
    \item \textbf{Step2-Parallel loading of KV caches:} We load the required new KV and the truncated reusable KV in parallel. The truncated excess data from the reusable KV is combined with the new required data, forming a consistent length of the new required KV.
    
    \item \textbf{Step3-Merge the data from all heads:} We combine the data from all heads (including both the new KV and the truncated reusable KV) to complete the KV reuse process. This allows efficient utilization of cached KV information, even with variable-length blocks.
\end{enumerate}

\textbf{The data in the abstract.} All data in the abstract are drawn from the main text or can be derived from it.



\end{document}